# The Rise and Fall of Robotic World (A case study of WALL-E)

Faisal Ghaffar
Systems Design Engineering
University of Waterloo
Waterloo, Canada
faisal.ghaffar@uwaterloo.ca

*Abstract*—The current trend in technology shows that the robots will soon be seen interacting with humans and handling different tasks more efficiently in future. With advancement in artificial intelligence it can be foreseen that robots will definitely take over all of the major and minor jobs, will socially interact with humans and will minimize the burden of human significantly. With the passage of time humans will forget about their reality and the labour work they currently do by themselves. That age will be exactly the peak age of robots and we can call it age of the rise of the robots. With the rise of robots, they will start taking decision by themselves regarding the human life and the planet. Human will again start to minimize the influence of robots and take control of their life with the help of some robots. The WALL·E narrates such a story of human life dependency on robots in spaceship when earth is destroyed because of some catastrophe. The clash between human and robots occurs when humans find the earth is survivable and should go back but robots do not allow them. With the help of some social robots humans again take control and take the ship back to earth. The major part of the movie is about the a lonely robot living on planet earth, the love of two robots, their interaction. Those two robots help humans to move back to earth.

*Index Terms*—Social robotics, Wall.e, future robots

## I. Introduction

This paper is based on the movie Wall-e [1] by Pixar Animation Studio which in turn is based on book wall-e [2]. The movie starts with showing a small intelligent robot named Wall-e which collects and arranges garbage on planet earth. The earth has been destroyed long ago because of some catastrophic event. To take shelter humans has shifted to somewhere else in the space and living on huge space aircraft. In that aircraft all the daily life activities are carried by robots.

The movie is 700 years in the future. The movie shows a city of skyscrapers made from compactly packed cube structure of trash. Wall.e robot is the last solar robot living human like life on earth. All of other such robots are also part of garbage as they have stopped working. Wall-e starts starts work at day time, it collects the garbage and then inside its body compresses it, change it to a cube shape and place it in organized manner on a pile. In the evening, Wall-E comes back to a large storage house, where he collected many items from the garbage and decorated them with Christmas lights. Wall-e turns on the television, where he watches human music and dance removes the belts from his wheels, and arrange some of the times, which he collected from the trash in his storage house and then goes into sleep mode. On the other side in the space air craft thousand of people are living and as all they activities are carried by robots, the humans have become lazy and fat.

It is lonely being WALL-E. But does WALL-E even know that? He comes home at night to a big storage area, where he has gathered a few treasures from his scavengings of the garbage and festooned them with Christmas lights. He wheels into his rest position, takes off his treads from his tired wheels and goes into sleep mode. Tomorrow is another day: One of thousands since the last humans left the Earth and settled into orbit aboard gigantic spaceships that resemble spas for the fat and lazy.

One day while working Wall-e came across a green plant, which is the first and only plant grown after ages on earth. Wall-e takes the plant to his storage and keeps it in a shoe. Meanwhile the spacecraft continuously sends mission to earth to find any signs of life. One day wall-e came across such mission in which he met a well-designed robot named Eva. Eva is quite aggressive, a mission oriented robot and is rigorously searching for something. Wall-e soon becomes friend with Eva after saving Eva from a dust storm. Wall-e takes Eva to his home and gives her things to entertain her. This is when Wall-e shows Eva the plant. When Eva sees the plant, it takes the plant, saves it with herself and signals the spacecraft that plant is found and then shuts down. Wall-e tries his best to wake up Eva but in vain. Soon the mission from space aircraft comes to earth and takes Eva with themselves. When Wall-e sees her he clings to the spacecraft to save Eva but ultimately ends up in spacecraft.

On spacecraft some protocol are then followed to collect the plant from Eva, but as the captain robot is in-charge and is in autopilot mode. It takes the decision by itself and without informing the human captain to destroy the plant as the captain robot assumes that earlier it is ordered to them, to not return to the earth. The human captain soon realized that they need to return to their homeland. He learns more about the earth and takes decision to return to earth but the autopilot does not allow it and takes the captain into custody. At this point Wall-e and Eva discovers the plant again and takes it to the captain himself. After a fight and showing some courage the human captain manages to stand on his feet, defeat the autopilot and turns off the auto-pilot mode. With the help of both Wall-e and Eva the captain takes control again and takes the ship back to earth and the life on earth starts once again.

## II. SOCIAL ASPECTS OF THE ROBOTS

### A. Roles of the robots and their place in human society

1) Wall-e: In the movie it is shown that earth has become a huge trash can where there are sky scrappers of garbage everywhere. It relates to the current world and wall-e is doing a trash collection job. Beside the garbage collection, Wall-e also compresses the garbage into cube structure and then organizes in the garbage city so that it occupies less space. All these works are currently carried out by humans, they do collect trash, using some machines they do compress it so that it occupies less space and if the garbage is recyclable, it is carried to a dumping place.

2) Eva: Eva is also a programmed robot whose job is scanning and searching. Eva is extraterrestrial vegetation evaluating robot, and its job is to find out whether there are some plants or not and hence find the truth whether earth is sustainable for human life or not. Eva also has the capability to sense danger and uses self protection protocol. In human society it can be easily related to search and rescue operation where humans search for an object and also use self protection protocols.

3) Misc robots: The miscellaneous robots are all those robots displayed in the movie. The scanning robots scan the foreign objects for contamination while the cleaning robots clean the spacecraft environment. The police robots acts as guards, they alert the people and also handles any kind of danger. Beside that the captain autopilot job is to look after all the activities going on in the spaceship and helps the captain in his decisions. All these robots carry out the tasks handled by humans on planet earth.

### B. Social skills of the robots

Throughout the movie social skills of all the robots have been exhibited. Some of those skills are below.

1) Communication: Both Wall-e and Eva robots seem to have non-verbal communication ability in the beginning. But later we found that they also communicate verbally. Wall-e robot initially interacts with a cockroach and extends its mechanical hand for asking to come with him [1, min.. 3:14]. Also in the movie when wall-e is saving Eva and boarding the shuttle plan wall-e uses mechanical hand signal to order the cockroach to stay back [1, min. 32:47]. Both Wall-e and Eva communicates with each other both verbally and also non-verbally. In the storage Wall-e gives different items to her and teaches how to use them. It exhibits the sign of happiness with dancing and sign of love by touching mechanical hands. It also interacts with humans on spacecraft to ask the human to move aside through its head and hand gestures.

2) Sharing: Wall-e shares its space with the cockroach in the movie and also offers its place to Eva. During his interaction with Eva, he also shares different kinds of items with Eva. At 01:15 in the movie when Wall-e threw the plant away, Wall-e grabs it again and reminds her that its important for her.

3) Coordination: The planet earth environment is alien to Eva, and it shoots every moving object. It doesn't know how to coordinate. But after some time it learns how to coordinate with the environment and other objects. When Wall-e extends its hand to give something to Eva it also extends its hand. When Wall-e and Eva travels to the spacecraft and they again lose the plant, both of them coordinate with each other to recover the plant. They signal each other when to hide and when to move and in which direction. At 51:00 and 1:08 in the movie Eva and Wall-e when interacting with captain, they keep the distance from the captain so they know their space well.

4) Empathy: Empathy is exhibited throughout the movie. Wall-e shares the same feeling about the plant as Eva when he knows about the importance.

## III. CAPABILITIES OF SYSTEM FROM TODAY'S PERSPECTIVE

### A. Behavior

1) Exploration behavior: Exploration behavior is the behavior while exploring a new environment. It basically consists of searching, attending, approaching and investigating steps [3]. Both Wall-e and Eva shows this behavior. Wall-e while searching through the trash examine and investigate different items while Eva scan the environment, search for a specific object and then analyze each object carefully. Brian et al. [4] discussed about frontier based explorations and how it can be extended to multiple robots so that they can coordinate and explore the space. In another study Wirth et al. [5] have also provided a robust solution for the exploration task of robots.

2) Appetitive behavior: Apetitive behavior is the activity that is goal oriented and it is carried out for satisfying a specific need, such as search for food. The Eva robot exhibits this behavior by searching restlessly for the plant while Wall-e robot shows this behavior by following Eva and searching for her. This behavior can be currently found in different robots such as iRoomba [6]. iRoomba robot shows appetitive behavior by searching for a trash or garbage.

3) Aversive behavior: Aversive behavior makes a robot capable to avoid any kind of obstacle, whether it is human, another object or an edge. The robot sensors the surrounding environment and acts accordingly. All of the robots in the movie shows aversive behavior. The robotic chairs which are carrying people in the movie avoid collision with each other. Even when an incident happens, all the robotic chairs halts its movement for a while until they are directed to follow an alternative path. The obstacle avoiding robot is developed using arduino [7], and it senses the environment and measures the distance to any object using its ultrasonic range finding sensor. Apolito et al. [8] developed a vision system for robots that enables a robot to avoid any kind of collisions

4) Path following behaviors: In earlier ages people used animals such as horses or donkeys for carrying their goods. These animals used to follow the path even in the absence of their owner. The same phenomenon can be seen in ants where they follow a designated path. In the movie this behavior can be observed by the cleaning robots and robotic chairs, They travel only on a designated line without moving away from it. This kind of behavior can be embedded in warehouse robots or assistive robots. Ma et al. [9] developed a conflict-free path planning problems for efficient guidance of multiple mobile robots. In a similar study Vivaldini et .al [10] also developed an algorithm that produces optimal routes for robots that are used in a warehouse for different kinds of works.

5) Postural behaviors: The upper body of wall-e is more like a humanoid and thus it has a human-like posture, the head moves in left and right direction as well as in upward and downward direction in the same way as humans. The lower body is mechonoid and uses wheels for locomotion. Overall wall-e has a great versatility and robustness in its posture. On the other hand Eva is a flapping-wing robot just like pigeonbot [11]. it soares in the air and does not touch the earth even in still position. Chang et al. developed such a bio-inspired robot and studied the aerodynamics of such robots and how they flip their wings and make maneuvers.

B. Decision making

Decision making is one of the main aspect of social robots. Decision making refers to how a robot responds in a certain situation. Dautenhahn et al. [10] describes the social robots as those robots which perceive and interpret the world in term of their own experience. Such capability can also be seen in the wall-e robot and it can also be deduced from various scenes in the movie that wall-e learns from the environment. When the Eva change some of the parts of wall-e, wall-e seems to have forgotten many of its skills and it doesn't recognize even wall-e [1, min. 1:26:30]. So it is clear that wall-e has learnt all the the social skills and and has been through a learning process. Some of the decision making skills of wall-e are as follows.

> Wall-e clearly understands the storm can harm him. when he sees a storm he decides to take shelter and runs towards his storage place [1, min. 8:19].
> wall-e decides which items to keep in storage and which items to compress in the garbage.
> He also decides where to keep a certain object in the rack [1, min. 07:19].
> When the solar power is low wall-e go up to the top of storage room for charging and when its charged, it then goes to do his tasks [1, min.. 09:50].

Similarly Eva robot also has an advanced level of decision making skills and it also acquires some of the skills during its time on the earth from wall-e. Initially the Eva shoots every moving object but after interaction with wall-e it then chooses which object to shoot and which not. Some of other decision making skills are below.

> The main job of Eva is to find whether earth environment is survivable or not and for this purpose it specifically looks for plants. It can clearly decide whether an object is a plant or not.
> Eva also decides where to hide wall-e while searching for the plant.
> Though the main job is to search the plant but when it finds the plant on spacecraft, it decides to throw away [1, min.. 1:42:50].
> Eva also searches for the right equipment for wall-e and decides which parts need to be changed [1, min.. 1:26:10].
> In interaction with the captain Eva decide whether to look at the captain or look at wall-e.

Without decision making skills, a robot might be unable to live in an environment with other people or objects. If a robot can decide when to stop or move it may collide with other objects in the environment. Shiomi et al. [12] develop a system that realizes human-like collision avoidance in a mobile robot to provide a comfortable environments for human. Currently the decision making is achieved with probabilistic models and training neural networks so that the robots become more independent and intelligent in making certain decisions.

IV. CURRENT TECHNOLOGY

The movie exhibits advancement of technology to such an extent that all the tasks are managed by robots. In this decade many technological advancements have been made and it has paved a path to make social robots which can handle different types of human work in better way than human. But still there are a lot of things such as emotions in which the robots lag behind. In the movie the captain autopilot can be seen to have an active argument with the human captain and takes control of space aircraft. With the current technology robots are not able to participate in such active arguments that leads to a serious war. The intelligence of robots with current technology is still at a very low level. Here we provide a brief comparison of the current available technology with the technology exhibited in the movie.

A. Software

With artificial intelligence and specifically deep learning it has become easier to design robots which can be easily integrated with the human environment. Some of the current software advancement, that can be embedded into robots, are discussed below.

1) Computer vision: Both wall-e and Eva robots use their fitted cameras as eyes and observe the environment. They use those cameras to understand the surrounding, differentiate between different types of objects and classify different items. Currently state of the art deep learning models can attain a classification level up to some extent. Some of the prominent models developed for classifications purposes are AlexNet [13], Inception [14], and ResNet [15]. Similarly many efficient algorithms have also been developed for object detection and object recognition. In a single image or scene multiple objects can be detected. The detection plays a vital role in scene understanding.

2) Speech processing: The robots in the movie understand speech, we can see that wall-e and Eva talk to each other in some scenes while it can also understand the broadcast announcement on the space aircraft. The current speech recog-nition technologies embedded as digital assistant devices such as Alexa and Siri easily understand and interpret the human speech. Apart from that many neural network models are also developed for speech denoising [16], multiple channel sepa-ration [17], music generation [18] music genre classification
[19] etc. These advancements match well with the capabilities of robots shown in the movie.

3) Learning process: The movie shows an advanced learn-ing process of the robots. They learn the complex phenomenon of the human world. The learning process shown in the movie cannot be replicated with current technology. Although with advancement of re-enforcement learning many people have developed many models that have defeated human experts, such as Alphago [20] developed by google deep-mind that defeated the human experts in the board game Go. Beside that Baker et al. [21] train multi agents using re-enforcement learning. These multi agents learn new strategies and exploit each other's weaknesses during their learning process.

4) Emotion exhibition: Throughout the movie the robots exhibit various emotions such as happy, angry, sad. Beside that Wall-e and Eva also develop a deep bond that can be compared to love between humans. Robots in the current era can interpret and understand different types of emotions. Usually they understand through reading human faces and respond with the matching emotions. RoboKind has created Milo [22] to help children with autism spectrum disorders learn more about emotional expression. Similarly Robin [23] was developed as a companion robot and its purpose was to give emotional support to children in hospital undergoing some kind of treatment. Though it is difficult to develop robots which can understand the deep chemistry with current technology we can still develop robots which can understand or exhibit basic emotions.

## B. Hardware

The hardware structure of wall-e is much more similar to the robots which are currently developed. The head structure comprises of two eyes (cameras) that helps in understanding and seeing the surrounding and objects. It is not clear how wall-e listens but it is quite obvious that listening can be duplicated with a microphone and its speaking capability with a speaker. For the hand movements and garbage compressing capability motors can be installed. Many modern robots such as Atlas can achieve this level of movement easily.

Eva has two extra capabilities, one is its flying ability and second is its firing ability. Beside these two abilities, other sensors are almost the same as wall-e. The pigeonbot
[11] duplicates its flying behavior while the Atlas robot can duplicate its firing behavior. The difference is that Atlas [24] robot need a gun to fire while the Eva robot has some builtin capability in its wings with which it fires.

The police robots can be seen with mounted screens on its front. These kind of hardware setup can be easily seen in commercially available robots nowadays. Moreover both Wall-e and Eva robots have touch sensor in their hands and also also they have motion sensors through which they can sense movement of other robots.

## V. CRITICAL DISCUSSION

The movie shows that human had live on space aircraft for almost 700 years and they completely rely on robots. Over reliance on technology can negatively affect human mental and physical growth and it is shown in a study conducted by Mesman et al. [25]. The same scenario can be shown in the movie. People do not talk to each other, neither are they aware of their surrounding but instead they are spending their time on robot chairs with screen to their front. The human became so lazy that they even cant sit on the chairs by themselves if they fall from the chair. This behavior shows that human has become so much dependent on technology that they have forgotten their real identity. Their mental and as well as physically health have been adversely affected by the technology. An advancement in technology which enhances the mental and physical health will be necessary, if the future is like this. The message of the film is clear that an end to human life can come because of over-reliance on technology.

One of the main themes in the movie is to showcase the dangers associated with giving a full autonomy to robots and letting them run all the human affairs. Giving a full autonomy to robots gives rise to many problems. The auto-pilot in the movie overrides the orders of the captain and that puts the human lives on stake. A human takes better precautions to save fellow human beings than a robot. Before giving autonomy to a robot it will be better to measure the trust level. Yagoda et al. [26] in their study developed a trust measurement tool based on performance, function and semantics. Beside that trust management mechanism should be developed for future robots and intelligent systems.

It is always thought that with advanced technology safety of a place can always be enhanced but there is always a limitation, and in many scenarios even the intelligent robots may fail. Despite the advanced technology and intelligent police robots on the spaceship, it fails to identify to wall-e intruder robot, even when it is roaming freely. It is because the the environment is not mutually run by human and robots but it is only run by robots.

The advanced technologies do not differentiate between a friend and a foe. If we carefully observe the behavior of Eva robot in the beginning, it fires to every moving object without analyzing and understanding the situation. This kind of behavior can cause serious damages not only to the envi-ronment but also to human life. When embedding a robot with such capabilities, an advanced decision making mechanism is necessary. Though the Eva robot has later learned that how and when to use this armour but still incidents can happen and one such incident happened on the spaceship when wall-e fires the same laser gun which Eva is using.

The strong emotion capability which wall-e learnt through the years might be beneficial in certain ways but for a goal-oriented and task specific robot might adversely affect the productivity of such robots and that may derail them from the job. Wall-e left its garbage collection task because of those emotions and wall-e also throws away the plant which is actually a critical mission for human survival. If the job is of critical nature it may again cause many problem in the environment in which it is working.